\documentclass[10pt,twocolumn,letterpaper]{article}

\usepackage{cvpr}
\usepackage{times}
\usepackage{epsfig}
\usepackage{graphicx}
\usepackage{amsmath}
\usepackage{amssymb}

\usepackage{gensymb}
\usepackage{mathptmx}
\usepackage{multirow}
\usepackage[caption=false]{subfig}
\usepackage[flushleft]{threeparttable}
\usepackage{rotating}
\usepackage[export]{adjustbox}
\usepackage{stackengine}

\usepackage[pagebackref=true,breaklinks=true,letterpaper=true,colorlinks,bookmarks=false]{hyperref}

\cvprfinalcopy % *** Uncomment this line for the final submission

\def\cvprPaperID{7} % *** Enter the CVPR Paper ID here

\ifcvprfinal\fi
\begin{document}

%%%%%%%%% TITLE
\title{Facial Expression Recognition from World \emph{Wild} Web}

%\author{First Author\\
%Institution1\\
%Institution1 address\\
%{\tt\small firstauthor@i1.org}
%% For a paper whose authors are all at the same institution,
%% omit the following lines up until the closing ``}''.
%% Additional authors and addresses can be added with ``\and'',
%% just like the second author.
%% To save space, use either the email address or home page, not both
%\and
%Second Author\\
%Institution2\\
%First line of institution2 address\\
%{\tt\small secondauthor@i2.org}
%}

\author{Ali Mollahosseini\textsuperscript{1}, Behzad Hassani\textsuperscript{1}, Michelle J. Salvador\textsuperscript{1}, \\Hojjat Abdollahi\textsuperscript{1}, David Chan\textsuperscript{2}, and Mohammad H. Mahoor\textsuperscript{1,2}\\
\textsuperscript{1} Department of Electrical and Computer Engineering\\
\textsuperscript{2} Department of Computer Science\\
University of Denver, Denver, CO\\
\tt\small{Ali.Mollahosseini@du.edu, Behzad.Hasani@du.edu, Michelle.Salvador@du.edu)}\\
\tt\small{habdolla@du.edu, davidchan@cs.du.edu, and mmahoor@du.edu}
}

\maketitle
\thispagestyle{empty}

%%%%%%%%% ABSTRACT
\begin{abstract}
Recognizing facial expression in a wild setting has remained a challenging task in computer vision. The World Wide Web is a good source of facial images which most of them are captured in uncontrolled conditions. In fact, the Internet is a Word Wild Web of facial images with expressions. This paper presents the results of a new study on collecting, annotating, and analyzing wild facial expressions from the web. Three search engines were queried using 1250 emotion related keywords in six different languages and the retrieved images were mapped by two annotators to six basic expressions and neutral. Deep neural networks and noise modeling were used in three different training scenarios to find how accurately facial expressions can be recognized when trained on noisy images collected from the web using query terms (e.g. happy face, laughing man, etc)? The results of our experiments show that deep neural networks can recognize wild facial expressions with an accuracy of 82.12\%.
 
\end{abstract}

%%%%%%%%% BODY TEXT
	\section{Introduction}

The World Wide Web (aka the Internet) has become a vast abundant source of information and data. Especially with the growth and use of social media and the availability of digital cameras on smart phones, people can easily add data to the Internet by taking photos, writing a short description, and immediately uploading them to the social media. People add more information to each photo by doing a tag, like, dislike, or comment on photos posted by friends or others on the Web. It is estimated that over 430 million photos are uploaded to Facebook and Instagram servers every day~\cite{instagramStat, facebookForbes}. Among photos posted on the Web, facial images have the highest incidents (e.g. selfies or self-portrait images are very popular nowadays). These facial photos are often taken in the wild under natural conditions with varying and diverse parameters such as scene lighting, user's head pose, camera view, image resolution and background, subject's gender, ethnicity, and facial expressions among others. Furthermore, the labels given by users use a wide range of vocabulary that is commonly understood to describe emotions, facial attributes, and expressions, of the pictures' contents. These photos are truly Wild images both in terms of the image quality/conditions and the labels given by users. An interesting question that may arise is, how the labels given wildly to facial images on the web by general users are consistent with the six basic emotions defined by psychologists.

On the other hand, computer vision and machine learning techniques for facial expression recognition are finding their ways into the design of a new generation of Human-Computer Interfaces. In order to train a machine learning system, many researchers have created databases using human actors/subjects portraying basic emotions~\cite{gross2010multiPie, pantic2005web, lyons1998coding}. However, most of the captured datasets mainly contain posed expressions acquired in a controlled environment. This is mostly due to the fact that it is hard and time consuming to collect unposed facial expression data in lab settings. However, in real applications, the system needs to capture and recognize spontaneous expressions, which involve different facial muscles, less exaggeration/intensity and have different dynamics than posed expressions. Researchers who have created spontaneous expression databases have captured the human face spontaneously while watching a short video or filling a questionnaires~\cite{grafsgaard2013automatically, mavadati2013disfa, mcduff2013affectiva}. However, the datasets are still captured in controlled lab settings (i.e. with the same illumination, resolution, etc.) or have a limited number of subjects, ethnicities, and poses poorly representing the environment and conditions faced in real-world situations. Existing databases in the wild settings, such as SFEW~\cite{dhall2011static} or FER2013~\cite{FER2013}, are also either very small or have low resolution without facial landmark points necessary for pre-processing.

Moreover, state-of-the-art machine learning algorithms such as Deep Neural Network requires big data for training and evaluation of the core algorithms. Given all the aforementioned motivations, this paper presents the results of our recent study with the aim of resolving the following questions:

\begin{enumerate}
	\item How consistent are the expression labels given by general web users compared to the six basic expression labels annotated by expert annotators on facial images?
	\item How accurately can a state-of-the-art algorithm classify images when trained on facial images collected from the web using query terms (e.g. happy face, laughing man, etc)?
\end{enumerate}

To address these questions, we created a database of in-the-wild facial expressions by querying different search engines (Google, Bing and Yahoo) We then annotated a subset of images using two human annotators and showed the general accuracy of the querying search engines for facial expression recognition. We trained two different deep neural network architectures with different training settings i.e. training on clean well-labeled data, training on a mixture of clean and noisy data, and training on mixture of clean and noisy data with a noise modeling approach using a general framework introduced in~\cite{xiao2015learning}. In other words, given the result of annotations, the noise level of each search engine is estimated as a prior distribution on the labels of our posterior set allowing for greater classification performance when we sample noisy labels and true labels in the same proportion. In order to achieve this, we learned a stochastic matrix where the entries are the probability of confusion in the labels. From this matrix, we can extract a posterior distribution on the true labels of the data conditioned on the true label given the noisy label, and the noisy label given the acquired data. For more information on the technique, see~\cite{xiao2015learning}.  

The rest of this paper is organized as follows. Section~\ref{sec:literature} reviews existing databases and state-of-the-art methods for facial expression recognition in the wild. Sec.~\ref{sec:ExpressionNet} explains the methodology of automatically collecting a large amount of facial expression images from the Internet and procedure of verifying them by two expert annotators. Section~\ref{sec:experimental} presents experimental results on training two different network architectures with different training settings, and section~\ref{sec:conclusion} concludes the paper. 

\section{Facial Expression Recognition in the wild}
\label{sec:literature}

Automatic Facial Expression Recognition (FER) is an important part of social interaction in Human-Machine-Interaction (HMI) systems~\cite{mollahosseini2014expressionbot}. Traditionally, automatic facial expression recognition (AFER) methods consist of three main steps – 1) registration and preprocessing, 2) feature extraction, and 3) classification. Preprocessing and registration form an important part of the AFER pipeline. Many studies have shown the advantages of using facial image registration to improve classification accuracy in both face identification and facial expression recognition~\cite{gritti2008local, rentzeperis2006impact}. In the feature extraction step, many methods such as HOG~\cite{mavadati2013disfa}, Gabor filters~\cite{liu2002gabor}, Local binary pattern (LBP)~\cite{shan2009facial}, facial landmarks~\cite{kobayashi1997facial}, pixel intensities~\cite{Mohammadi2014PCA_based}, and Local phase quantization (LPQ)~\cite{zhen2012facial}, or a combination of multiple features using multiple kernel learning methods~\cite{zhang2015facial, zhang2014ebear} have been proposed to extract discriminative features. Classification is the final step of most AFER techniques. Support vector machines~\cite{zhen2012facial}, multiple kernel learning~\cite{zhang2015facial, zhang2014ebear}, dictionary learning~\cite{mohammadi2015intensity} etc. have been shown to have a great performance in classifying discriminative features extracted from the previous stage.

Although, traditional machine learning approaches have been successful when classifying posed facial expressions in a controlled environment, they do not have the flexibility to classify images captured in a spontaneous uncontrolled manner (``in the wild'') or when applied to databases for which they were not designed. The poor generalizability of traditional methods is primarily due to the fact that many approaches are subject or database dependent and only capable of recognizing exaggerated or limited expressions similar to those in the training database. Many FER databases have tightly controlled illumination and pose conditions. In addition, obtaining accurate training data is particularly difficult, especially for emotions such as sadness or fear which are extremely difficult to accurately replicate and do not occur often in real life.

\begin{figure}
\centering
\subfloat
{
	\centering
	\stackunder{\includegraphics[width=20mm]{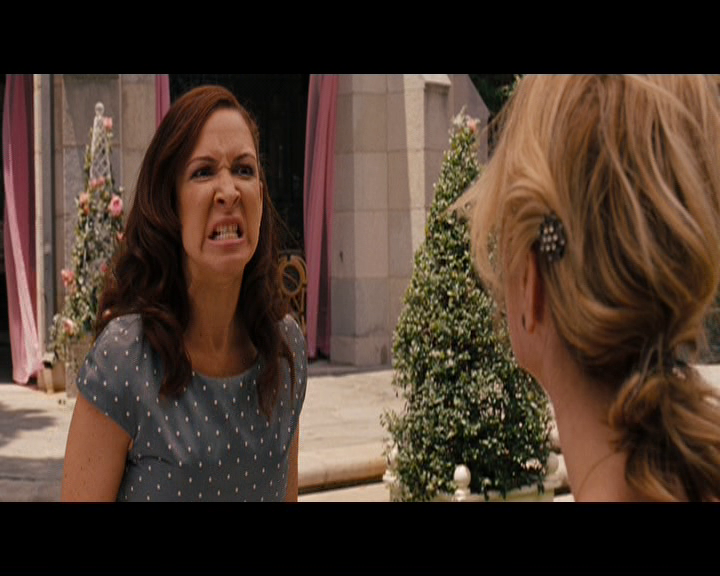}}{\scriptsize{Angry}}%
}
\subfloat
{
	\centering
	\stackunder{\includegraphics[width=20mm]{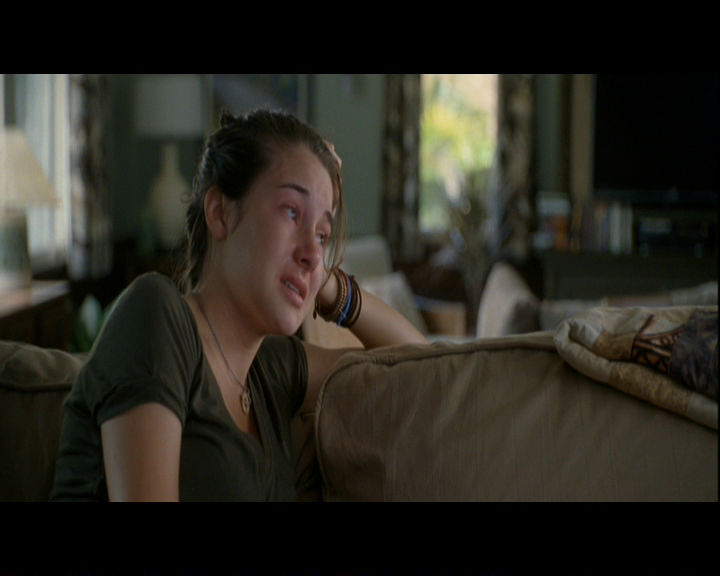}}{\scriptsize{Sad}}%
}
\subfloat
{
	\centering
	\stackunder{\includegraphics[width=20mm]{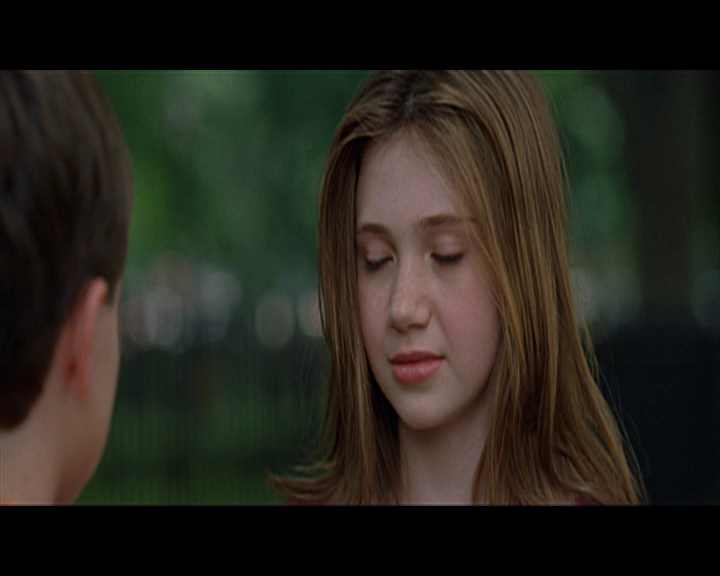}}{\scriptsize{Disgust}}%
}
\subfloat
{
	\centering
	\stackunder{\includegraphics[width=20mm]{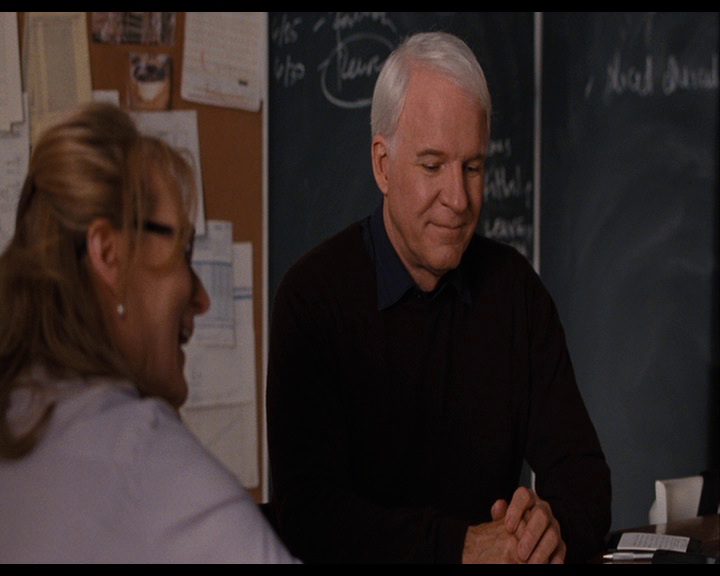}}{\scriptsize{Happy}}%
}
\vspace{-0.2cm}
\subfloat
{
	\centering
	\stackunder{\includegraphics[width=20mm,height=20mm]{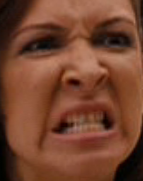}}{\scriptsize{}}%
}
\subfloat
{
	\centering
	\stackunder{\includegraphics[width=20mm,height=20mm]{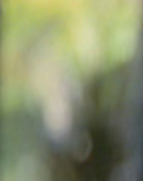}}{\scriptsize{}}%
}
\subfloat
{
	\centering
	\stackunder{\includegraphics[width=20mm,height=20mm]{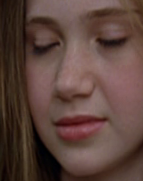}}{\scriptsize{}}%
}
\subfloat
{
	\centering
	\stackunder{\includegraphics[width=20mm,height=20mm]{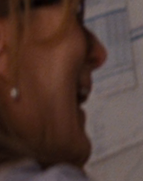}}{\scriptsize{}}%
}
\caption{\label{fig:Sample_SFEW}
Sample of images from SFEW~\cite{dhall2011static} and their original registered images published with the database. 
}
\end{figure}

Recently, facial expression datasets with in the wild settings have attracted much attention. Dhall~\emph{et al.}~\cite{dhall2013emotion} released Acted \textit{Facial Expressions in the Wild (AFEW)} from movies by semi-automatic approach via a recommender system based on subtitles. AFEW addresses the issue of temporal facial expressions and it is the only temporal publicly available facial expression database in the wild. A static subset \textit{Static Facial Expressions in the Wild (SFEW)} is created by selecting static frames which covers unconstrained facial expressions, different head poses, age range, and occlusions and close to real world illuminations. However, it contains only 1635 images and there are only 95 subjects in the database. In addition, due to the wild settings of the database, the released facial location and landmarks do not capture the faces in all images correctly making some training and test samples unusable (See Fig.~\ref{fig:Sample_SFEW}). 

\begin{figure}
\centering
\subfloat
{
	\centering
	\stackunder{\includegraphics[width=11mm]{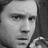}}{\scriptsize{}}%
}
\subfloat
{
	\centering
	\stackunder{\includegraphics[width=11mm]{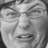}}{\scriptsize{}}%
}
\subfloat
{
	\centering
	\stackunder{\includegraphics[width=11mm]{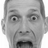}}{\scriptsize{}}%
}
\subfloat
{
	\centering
	\stackunder{\includegraphics[width=11mm]{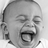}}{\scriptsize{}}%
}
\subfloat
{
	\centering
	\stackunder{\includegraphics[width=11mm]{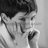}}{\scriptsize{}}%
}
\subfloat
{
	\centering
	\stackunder{\includegraphics[width=11mm]{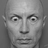}}{\scriptsize{}}%
}
\subfloat
{
	\centering
	\stackunder{\includegraphics[width=11mm]{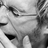}}{\scriptsize{}}%
}

\vspace{-0.3cm}
\subfloat
{
	\centering
	\stackunder{\includegraphics[width=11mm]{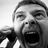}}{\scriptsize{Angry}}%
}
\subfloat
{
	\centering
	\stackunder{\includegraphics[width=11mm]{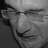}}{\scriptsize{Disgust}}%
}
\subfloat
{
	\centering
	\stackunder{\includegraphics[width=11mm]{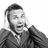}}{\scriptsize{Fear}}%
}
\subfloat
{
	\centering
	\stackunder{\includegraphics[width=11mm]{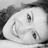}}{\scriptsize{Happy}}%
}
\subfloat
{
	\centering
	\stackunder{\includegraphics[width=11mm]{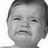}}{\scriptsize{Sad}}%
}
\subfloat
{
	\centering
	\stackunder{\includegraphics[width=11mm]{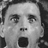}}{\scriptsize{Surprise}}%
}
\subfloat
{
	\centering
	\stackunder{\includegraphics[width=11mm]{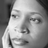}}{\scriptsize{Neutral}}%
}
\vspace{0.3cm}

\caption{\label{fig:Sample_FER2013}
Sample of images from FER2013 database~\cite{FER2013}. 
}
\end{figure}

The Facial Expression Recognition 2013 (FER-2013) database was introduced in the ICML 2013 Challenges in Representation Learning~\cite{FER2013}. The database was created using the Google image search API that match a set of 184 emotion-related keywords to capture the six basic expressions as well as the neutral expression. Human labelers rejected incorrectly labeled images. Images are resized to 48x48 pixels and converted to grayscale. The resulting database contains 35,887 images most of them in wild settings, yet only 547 of the images portray disgust. Figure~\ref{fig:Sample_FER2013} shows some sample images of FER2013. FER2013 is currently the biggest publicly available facial expression database in wild settings, enabling many researchers to train machine learning methods where large amounts of data are needed such as Deep neural networks. However, as shown in Fig.~\ref{fig:Sample_FER2013}, the faces are not registered, and unfortunately most of facial landmark detectors fail to extract facial landmarks at this resolution and quality. 

In addition, FER in the wild is really a challenging task both in terms of machine and human performance. Extensive experiments in~\cite{gehrig2013facial} show that even humans are only capable of 53\% agreement in terms of Fleiss’ kappa over all classes to classify AFEW video clips without listening to the audio track. State-of-the-art automated methods have achieved 35\% accuracy on AFEW video clips by using audio modalities~\cite{zong2016emotion}. Even recognizing expression from still images or static frames using traditional machine learning approaches are not accurate and the best performance on SFEW 2.0 database is reported as 50\% accuracy (with a baseline of 39.13\%) ~\cite{zong2016emotion}. 

Recently, deep neural networks have seen a resurgence in popularity. Recent state-of-the-art results have been obtained using neural networks in the fields of visual object recognition~\cite{krizhevsky2012imagenet, szegedy2014going}, human pose estimation~\cite{toshev2014deeppose}, face verification~\cite{taigman2014deepface}, and many more. Even in the FER field results so far have been promising~\cite{kahou2013combining, mollahosseini2015going, liu2013aware, kahou2013combining}, and most of the facial expression recognition challenge winners have used deep neural networks~\cite{tang2013deep, yu2015image}.

In the FER problem, however, unlike visual object databases such as imageNet~\cite{deng2009imagenet}, existing FER databases often have limited numbers of subjects, few sample images or videos per expression, or small variation between sets, making neural networks significantly more difficult to train. For example, the FER2013 database~\cite{FER2013} (one of the largest recently released FER databases) contains 35,887 images of different subjects yet only less than 2\% of the images portray disgust. Similarly, the CMU MultiPIE face database~\cite{gross2010multiPie} contains around 750,000 images but is comprised of only 337 different subjects, where 348,000 images portray only a ``neutral'' emotion and the remaining images do not portray anger, fear or sadness.

In a  recent study~\cite{mollahosseini2015going}, the authors proposed a deep neural network architecture and combined seven well-known facial expression databases (i.e. MultiPIE, MMI, CK+, DISFA, FERA, SFEW, and FER2013) to perform an extensive study on subject-independent and cross database. The results of the proposed architecture were comparable to or better than the state-of-the-art methods, However, the majority of data were still posed images and performance on wild databases (SFEW and FER2013) were only comparable to the state-of-the-art methods. 

Considering the need to develop an automated FER in wild system, and issues with the current facial expression in wild databases, a possible solution is to automatically collect a large amount of facial expression images from the abundant images available on the Internet, and directly use them as ground truth to train deep models. However, consideration should be done to avoid false samples in the search engine results for expressions such as disgust or fear. This is due to the higher tendancy of people to publish happy or neutral faces that can be mislabeled or associated with disgust or fear by web users. 

Nonetheless, semi-supervised~\cite{weston2012deep}, transfer learning~\cite{oquab2014learning}, or noise modeling approaches~\cite{sukhbaatar2014learning, xiao2015learning} can be used to train deep neural networks with noisy data by obtaining large amounts of facial expression images from search engines, along with a smaller subset of fully well-labeled images.  

% % % % % % % % % % % % % %
%-------------------------------------------------------------------------

\section{Facial expressions from the \textit{wild} web}
\label{sec:ExpressionNet}

To create our database with the larger amount of images necessary for Deep Neural Networks, three search engines were queried by facial emotion related tags in six different languages. We used Google, Bing, and Yahoo. Other search engines were considered such as Baidu and Yandex. However they either did not produce a high percentage of the intended images or they did not have accessible APIs for automatically querying and pulling image urls into the database.

A total of 1250 search queries were compiled in six languages and used to crawl Internet search engines for the image urls in our dataset. The first 200 urls returned for each query were stored in the database (258,140 distinct urls). Among the 258,140 urls, 201,932 images were available for download. OpenCV face recognition was used to obtain bounding boxes around each face. Bidirectional warping of Active Appearance Model (AAM)~\cite{mollahosseini2013bidirectional} and a face alignment algorithm via regressing local binary features~\cite{ren2014face, LequanYu2016} were used to extract 66 facial landmarks. The employed facial landmark localization techniques have been trained using the annotations provided from the 300W competition~\cite{sagonas2015300, sagonas2013semi, sagonas2013300}. Images with at least one face with facial landmark points were kept for the next processing stages. A total of 119,481 images were kept. Other attributes of the queries were stored if applicable such as; intended emotion, gender, age, language searched, and its English translation if not in English.

\begin{figure}
\centering
\subfloat
{
	\centering
	\stackunder{\includegraphics[width=18mm]{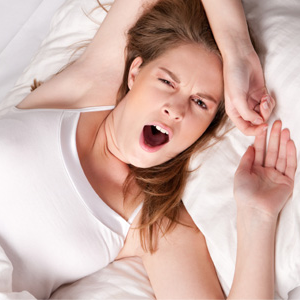}}{\scriptsize{\emph{None} (Sad)}}%
}
\subfloat
{
	\centering
	\stackunder{\includegraphics[width=18mm]{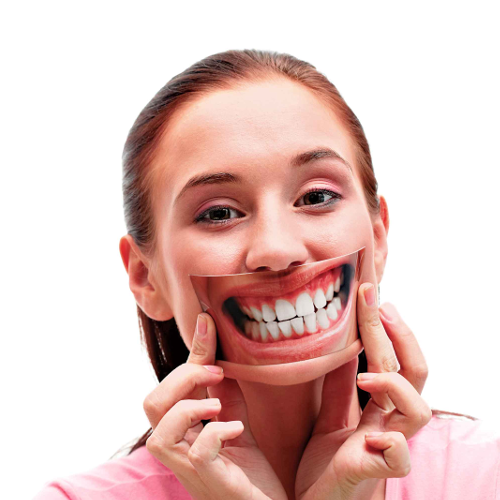}}{\scriptsize{\emph{No-Face} (Happy)}}%
}
\subfloat
{
	\centering
	\stackunder{\includegraphics[width=18mm]{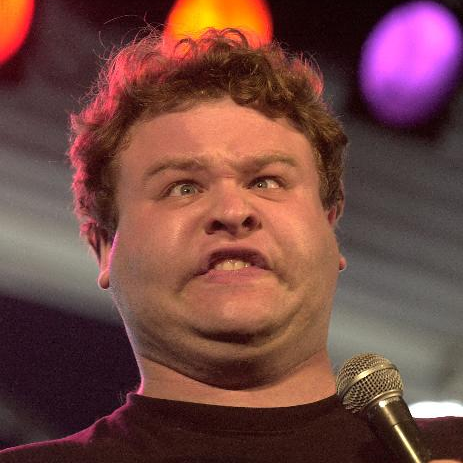}}{\scriptsize{\emph{Uncertain} (Angry)}}%
}
\subfloat
{
	\centering
	\stackunder{\includegraphics[width=18mm]{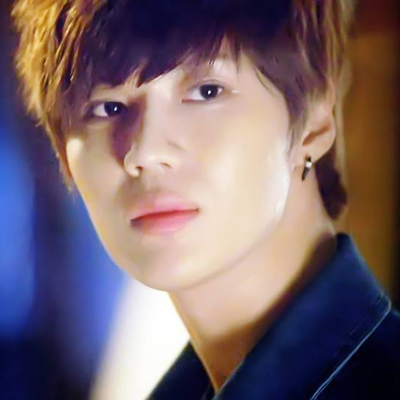}}{\scriptsize{\emph{Neutral} (Disgust)}}%
}
\vspace{-0.2cm}
\subfloat
{
	\centering
	\stackunder{\includegraphics[width=18mm]{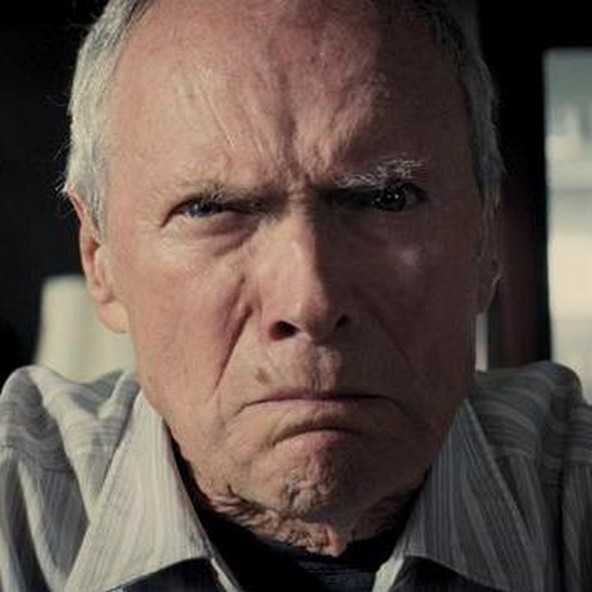}}{\scriptsize{\emph{Angry} (Happy)}}%
}
\subfloat
{
	\centering
	\stackunder{\includegraphics[width=18mm]{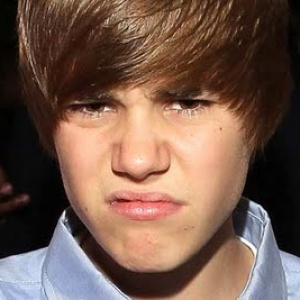}}{\scriptsize{\emph{Disgust} (Angry)}}%
}
\subfloat
{
	\centering
	\stackunder{\includegraphics[width=18mm]{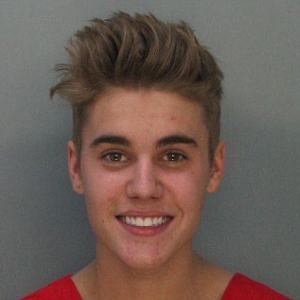}}{\scriptsize{\emph{Happy} (Angry)}}%
}
\subfloat
{
	\centering
	\stackunder{\includegraphics[width=18mm]{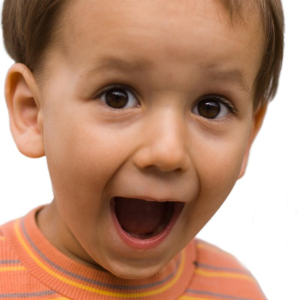}}{\scriptsize{\emph{Surprise} (Happy)}}%
}
\vspace{-0.2cm}
\subfloat
{
	\centering
	\stackunder{\includegraphics[width=18mm]{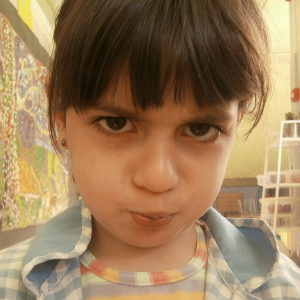}}{\scriptsize{\emph{Angry} (Angry)}}%
}
\subfloat
{
	\centering
	\stackunder{\includegraphics[width=18mm]{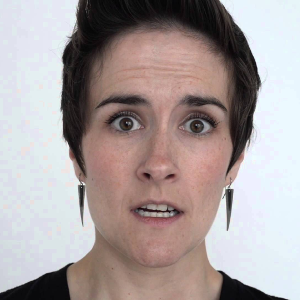}}{\scriptsize{\emph{Fear} (Fear)}}%
}
\subfloat
{
	\centering
	\stackunder{\includegraphics[width=18mm]{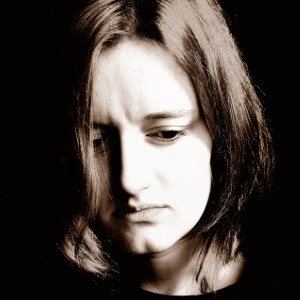}}{\scriptsize{\emph{Sad} (Sad)}}%
}
\subfloat
{
	\centering
	\stackunder{\includegraphics[width=18mm]{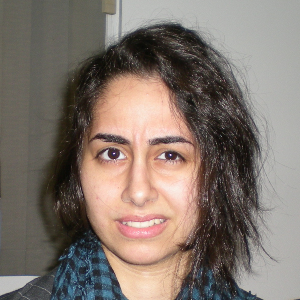}}{\scriptsize{\emph{Disgust} (Disgust)}}%
}
\vspace{0.3cm}
\caption{\label{fig:Sample_ExpressionNet}
Sample of queried images from the web and their annotated tags. The queried expression is written in parentheses.
}
\end{figure}

On average 4000 images of each queried emotions were selected randomly, and in total 24,000 images were given to two expert annotators to categorize the face in the image into nine categories (i.e. No-face, six basic expressions, Neutral, None, and Uncertain). The annotators were instructed to select the proper expression category on the face, where the intensity is not important as long as the face depicts the intended expressions. The \emph{No-face} category was defined as images that: 1) There was no face in the image; 2) There was a watermark on the face; 3) The bounding box was not on the face or did not cover the majority of the face; 3) The face is a drawing, animation, painted, or printed on something else; and 4) The face is distorted beyond a natural or normal shape, even if an expression could be inferred. The \emph{None} category was defined as images that portrayed an emotion but the expression/emotions could be categorized as one of the six basic emotions or neutral (such as sleepy, bored, tired, seducing, confused, shame, focused,  etc.). If the annotators were uncertain about any of the facial expressions, images were tagged as \emph{uncertain}. Figure~\ref*{fig:Sample_ExpressionNet} shows some examples of each category and the intended queries written in parentheses.

The annotation was performed fully blind and independently, i.e. the annotators were not aware of the intended query or other annotator's response. The two annotators agreed on 63.7\% of the images. For the images that were at a disagreement, favor was given to the intended query i.e. if one of the annotators labeled the image as the intended query, the image was labeled in the database with the intended query. This happened in 29.5\% of the images with disagreement between the annotators. On the rest of the images with disagreement, one of the annotations was assigned to the image randomly. Table~\ref{Tab:NumImages} shows the number of images in each category in the set of 24,000 images that were given to two human annotators. As shown, some expressions such as Disgust, Fear, and Surprise have few images compared to the other expressions, despite the number of queries being the same.

\begin{table}[]
	\centering
	\caption{Number of annotated images in each category}
	\label{Tab:NumImages}
	\begin{tabular}{|l|c|}
		\hline
		\textbf{Label} & \multicolumn{1}{l|}{\textbf{Number of images}} \\ \hline
		Neutral        & 3501                                           \\ \hline
		Happy          & 7130                                           \\ \hline
		Sad            & 3128                                           \\ \hline
		Surprise       & 1439                                           \\ \hline
		Fear           & 1307                                           \\ \hline
		Disgust        & 702                                            \\ \hline
		Anger          & 2355                                           \\ \hline
		None           & 403                                            \\ \hline
		Uncertain      & 280                                            \\ \hline
		No-face        & 3755                                           \\ \hline
	\end{tabular}
\end{table}

Table~\ref{Tab:annotationConfMatrix} shows the confusion matrix between queried emotions and their annotations. As is shown, happiness had the highest hit-rate (68\%) and the rest of emotions had hit-rates at less than 50\%. There was about 15\% confusion with \emph{No-Face} category for all emotions, as many images from the web contained watermarks, drawings etc. About 15\% of all queried emotions resulted in neutral faces. Disgust and Fear had the lowest hit rate among other expression with 12\% and 17\% hit-rates respectively and most of the result of disgust and fear are mainly happiness or \emph{No-Face}.

\begin{table*}[t]
\centering
\caption{Confusion Matrix of annotated images for different intended emotion-related query terms}
\label{Tab:annotationConfMatrix}
\begin{tabular}{l|c|c|c|c|c|c|c|c|c|c|}
\cline{2-11}
                                        & \textbf{Happy} & \textbf{Sad}   & \textbf{Surprise} & \textbf{Fear} & \textbf{Disgust} & \textbf{Anger} & \textbf{Neutral} & \textbf{No-Face} & \textbf{None} & \textbf{Uncertain} \\ \hline
\multicolumn{1}{|l|}{\textbf{Happy}}    & \textbf{68.18} & 2.66           & 1.23              & 0.74          & 0.33             & 1.59           & 5.67             & 18.54            & 0.74          & 0.33               \\ \hline
\multicolumn{1}{|l|}{\textbf{Sad}}      & 16.5           & \textbf{42.42} & 1.52              & 1.88          & 0.57             & 4.73           & 16.55            & 13.31            & 1.57          & 0.98               \\ \hline
\multicolumn{1}{|l|}{\textbf{Surprise}} & \textbf{27.6}           & 6.31           & 20.11    & 5.62          & 1.07             & 4.85           & 17.1             & 14.73            & 1.65          & 0.96               \\ \hline
\multicolumn{1}{|l|}{\textbf{Fear}}     & 18.74          & 10.91          & 6.49              & 17.69         & 1.47             & 6.39           & 13.92            & \textbf{20.49}   & 2.22          & 1.67               \\ \hline
\multicolumn{1}{|l|}{\textbf{Disgust}}  & \textbf{26.71} & 7.47           & 4.48              & 4.53          & 12.61            & 9.62           & 17.34            & 12.41            & 2.99          & 1.84               \\ \hline
\multicolumn{1}{|l|}{\textbf{Anger}}    & 22.28          & 7.39           & 2.31              & 2.11          & 1.19             & \textbf{30.59} & 16.21            & 14.43            & 2.34          & 1.14               \\ \hline
\end{tabular}
\end{table*}

\section{Training from web-images}
\label{sec:experimental}

The annotated images labeled with six basic expressions as well as neutral faces are selected from 24,000 annotated images (18,674 images). Twenty percent of each label is randomly selected as a test set (2,926 images) and the rest are used as training and validation sets. A total of 60K of not annotated images (10K for each basic emotion) is selected as noisy training set.    

As baselines, two different deep neural network architectures are trained in three different training scenarios: 1) training on well-labeled images, 2) training on a mixture of noisy and well-labeled sets, and 3) training on a mixture of noisy and well-labeled sets using a noise modeling approach introduced in~\cite{xiao2015learning}. The network architecture we used in these experiments are AlexNet~\cite{krizhevsky2012imagenet} and 
a networks for facial expression recognition recently published in WACV2016 in~\cite{mollahosseini2015going}, called WACV-Net in the rest of this paper. All networks are evaluated on a well-labeled test set. 

AlexNet consists of five convolutional layers, some of which are followed by max-pooling layers, and three fully-connected layers. To augment the data, ten crops of registered facial images of size 227x227 pixels are fed to AlexNet. We have tried a smaller version of AlexNet with smaller input images of 40x40 pixels and smaller convolutional kernel sizes, but the results were not as promising as the original model. WACV-Net consists of two convolutional layers each followed by max pooling, four Inception layers, and two fully-connected layers. The input images are resized to 48x48 pixels with ten augmented crops of 40x40 pixels. Our version of AlexNet performed more than 100M operations, whereas the WACV-Net performs about 25M operations, due to size reductions in Inception layers. Therefore, WACV-Net trained almost four times faster than AlexNet and consequently it had faster evaluation time as well.  

In the first scenario, the network is trained on only well-labeled set with random initialization. In the second scenario (mixture of noisy and well-labeled sets), the network is pre-trained with only well-labeled data, and then trained on the mixture of the noisy and well-labeled sets. This increased about 5\% in accuracy compared with training on the mixture of the noisy and well-labeled sets from scratch. In the last scenario (mixture of noisy and well-labeled sets using the noise modeling), as the posterior computation could be totally wrong if the network is randomly initialized~\cite{xiao2015learning}, the network components are pre-trained with the well-labeled data. In addition, we bootstrap/upsample the well-labeled data to half of the noisy data. In all scenarios, we used a mini-batch size of 256. The learning rate is initialized to be 0.001 and is divided by 10 after every 10,000 iterations. We keep training each model until convergence.

Table~\ref{Tab:RecognitionAccuracy} shows the overall recognition accuracy of AlexNet and WACV-Net on the test set in three training scenarios. As shown, in all cases AlexNet performed better than WACV-Net. Training on mixture of the noisy and well-labeled data were not as successful as training on only well-labeled data. We believe that this was due to the fact that facial expression images crawled from the web are very noisy and in most expressions, less than 50\% of the noisy data portray the intended query. The noise estimation approach can improve the accuracy of the network trained on the mixture of noisy and well-labeled sets. The best result is achieved from training AlexNet on well-labeled data. This gives slightly better overall accuracy (1\%) than training on the mixture of noisy and well-labeled sets using noise modeling.

\begin{table}[]
	\centering
	\caption{Recognition accuracy of AlexNet and WACV-Net on well-labeled test set with different training settings}
	\label{Tab:RecognitionAccuracy}
	\begin{tabular}{l|l|l|}
		\cline{2-3}
		& AlexNet & WACV-Net~\cite{mollahosseini2015going} \\ \hline
		\multicolumn{1}{|l|}{Train on well-labeled}                                   &  \multicolumn{1}{c|}{82.12\%}  &  \multicolumn{1}{c|}{75.15\%}    \\ \hline
		\multicolumn{1}{|l|}{Train on mix}                                            &  \multicolumn{1}{c|}{69.03\%}  &  \multicolumn{1}{c|}{67.04\%}    \\ \hline
		\multicolumn{1}{|l|}{\begin{tabular}{@{}l@{}} Train on mix with \\noise estimation~\cite{xiao2015learning}
		\end{tabular}} &  \multicolumn{1}{c|}{81.68\%}  &  \multicolumn{1}{c|}{76.52\%}    \\ \hline
	\end{tabular}
\end{table}

Table~\ref{Tab:Confusion_AlexNet_case1} shows the confusion matrix of AlexNet trained on the well-labeled set. Table~\ref{Tab:Confusion_AlexNet_case3} shows the confusion matrix of AlexNet trained on the mixture of noisy and well-labeled sets with noise estimation~\cite{xiao2015learning}. As shown in these tables, the noise estimation approach can improve the recognition accuracy of sadness, surprise, fear and disgust expressions. The reason is that there are fewer samples of these expressions in the well-labeled sets compared with other labels, and therefore including noisy data increases the training samples if the posterior distribution is estimated well. However, in some cases such as neutral faces and angry, training on only well-labeled data has higher recognition accuracy, as the prior distribution on the well-label set may not fully reflect the posterior distribution on the noisy set. 
  
\begin{table}[]
	\caption{Confusion matrix of AlexNet Trained on well-labeled}
	\label{Tab:Confusion_AlexNet_case1}
	\centering
	\begin{tabular}{cc|c|c|c|c|c|c|c|}
		\cline{3-9}
		&             & \multicolumn{7}{c|}{\scriptsize{predicted}}                                                                  \\ \cline{3-9} 
		&             & \textbf{\scriptsize{NE}} & \textbf{\scriptsize{HA}} & \textbf{\scriptsize{SA}} & \textbf{\scriptsize{SU}} & \textbf{\scriptsize{FE}} & \textbf{\scriptsize{DI}} & \textbf{\scriptsize{AN}} \\ \hline
		\multicolumn{1}{|c|}{\multirow{7}{*}{\begin{turn}{90}\scriptsize{Actual}\end{turn}
			}} & \textbf{\scriptsize{NE}} & \scriptsize{\textbf{79.12}}        & \scriptsize{6.73}        & \scriptsize{9.98}        & \scriptsize{0.46}         & \scriptsize{0}        & \scriptsize{0}        & \scriptsize{3.71}         \\ \cline{2-9} 
			\multicolumn{1}{|c|}{}                        & \textbf{\scriptsize{HA}} & \scriptsize{6.37}        & \scriptsize{\textbf{91.63}}        & \scriptsize{1.14}         & \scriptsize{0.29}       & \scriptsize{0.14}        & \scriptsize{0.07}        & \scriptsize{0.36}         \\ \cline{2-9} 
			\multicolumn{1}{|c|}{}                        & \textbf{\scriptsize{SA}} & \scriptsize{14.52}         & \scriptsize{5.24}         & \scriptsize{\textbf{73.10}}        & \scriptsize{0.24}        & \scriptsize{0.48}        & \scriptsize{0.71}         & \scriptsize{5.71}        \\ \cline{2-9} 
			\multicolumn{1}{|c|}{}                        & \textbf{\scriptsize{SU}} & \scriptsize{10.59}         & \scriptsize{6.47}         & \scriptsize{1.18}         & \scriptsize{\textbf{76.47}}        & \scriptsize{3.53}         & \scriptsize{1.18}         & \scriptsize{0.59}         \\ \cline{2-9} 
			\multicolumn{1}{|c|}{}                        & \textbf{\scriptsize{FE}} & \scriptsize{4.14}         & \scriptsize{3.45}         & \scriptsize{7.59}         & \scriptsize{15.86}        & \scriptsize{\textbf{60}}        & \scriptsize{2.76}         & \scriptsize{6.21}         \\ \cline{2-9} 
			\multicolumn{1}{|c|}{}                        & \textbf{\scriptsize{DI}} & \scriptsize{2.41}           & \scriptsize{4.82}         & \scriptsize{8.43}        & \scriptsize{2.41}         & \scriptsize{1.2}        & \scriptsize{\textbf{57.83}}        & \scriptsize{22.89}         \\ \cline{2-9} 
			\multicolumn{1}{|c|}{}                        & \textbf{\scriptsize{AN}} & \scriptsize{8.6}         & \scriptsize{2.87}         & \scriptsize{5.73}        & \scriptsize{1.79}        & \scriptsize{0.36}        & \scriptsize{5.73}         & \scriptsize{\textbf{74.91}}        \\ \hline
	\end{tabular}
	 \begin{tablenotes}
	    	\footnotesize
	    	\item \textsuperscript{*} NE, HA, SA, SU, FE, DI, AN stand for Neutral, Happiness, Sadness, Surprised, Fear, Disgust, Anger respectively.        
	 \end{tablenotes}	
\end{table}	

\begin{table}[]
	\caption{Confusion matrix of AlexNet Trained mixture of noisy and well-labeled sets with noise estimation}
	\label{Tab:Confusion_AlexNet_case3}
	\centering
	\begin{tabular}{cc|c|c|c|c|c|c|c|}
		\cline{3-9}
		&             & \multicolumn{7}{c|}{\scriptsize{predicted}}                                                                  \\ \cline{3-9} 
		&             & \textbf{\scriptsize{NE}} & \textbf{\scriptsize{HA}} & \textbf{\scriptsize{SA}} & \textbf{\scriptsize{SU}} & \textbf{\scriptsize{FE}} & \textbf{\scriptsize{DI}} & \textbf{\scriptsize{AN}} \\ \hline
		\multicolumn{1}{|c|}{\multirow{7}{*}{\begin{turn}{90}\scriptsize{Actual}\end{turn}
			}} & \textbf{\scriptsize{NE}} & \scriptsize{\textbf{65.20}}        & \scriptsize{10.67}        & \scriptsize{20.19}        & \scriptsize{0.23}         & \scriptsize{0}        & \scriptsize{1.62}        & \scriptsize{2.09}         \\ \cline{2-9} 
			\multicolumn{1}{|c|}{}                        & \textbf{\scriptsize{HA}} & \scriptsize{3.29}        & \scriptsize{\textbf{91.56}}        & \scriptsize{3.72}         & \scriptsize{0.21}       & \scriptsize{0.21}        & \scriptsize{0.43}        & \scriptsize{0.57}         \\ \cline{2-9} 
			\multicolumn{1}{|c|}{}                        & \textbf{\scriptsize{SA}} & \scriptsize{7.62}         & \scriptsize{3.33}         & \scriptsize{\textbf{84.29}}        & \scriptsize{0.24}        & \scriptsize{0.95}        & \scriptsize{1.43}         & \scriptsize{2.14}        \\ \cline{2-9} 
			\multicolumn{1}{|c|}{}                        & \textbf{\scriptsize{SU}} & \scriptsize{5.29}         & \scriptsize{5.88}         & \scriptsize{4.12}         & \scriptsize{\textbf{76.47}}        & \scriptsize{5.29}         & \scriptsize{1.76}         & \scriptsize{1.18}         \\ \cline{2-9} 
			\multicolumn{1}{|c|}{}                        & \textbf{\scriptsize{FE}} & \scriptsize{0.69}         & \scriptsize{3.45}         & \scriptsize{11.72}         & \scriptsize{13.79}        & \scriptsize{\textbf{63.45}}        & \scriptsize{3.45}         & \scriptsize{3.45}         \\ \cline{2-9} 
			\multicolumn{1}{|c|}{}                        & \textbf{\scriptsize{DI}} & \scriptsize{2.41}           & \scriptsize{4.82}         & \scriptsize{6.02}        & \scriptsize{2.41}         & \scriptsize{1.2}        & \scriptsize{\textbf{68.67}}        & \scriptsize{14.46}         \\ \cline{2-9} 
			\multicolumn{1}{|c|}{}                        & \textbf{\scriptsize{AN}} & \scriptsize{6.45}         & \scriptsize{2.87}         & \scriptsize{12.54}        & \scriptsize{2.87}        & \scriptsize{0.36}        & \scriptsize{4.66}         & \scriptsize{\textbf{70.25}}        \\ \hline
	\end{tabular}
	 \begin{tablenotes}
	    	\footnotesize
	    	\item \textsuperscript{*} NE, HA, SA, SU, FE, DI, AN stand for Neutral, Happiness, Sadness, Surprised, Fear, Disgust, Anger respectively.        
	 \end{tablenotes}	
\end{table}	
	
\begin{figure}
\centering
\subfloat
{
	\centering
	\stackunder{\includegraphics[width=18mm]{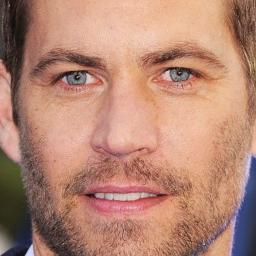}}{\scriptsize{\emph{Happy} (Neutral)}}%
}
\subfloat
{
	\centering
	\stackunder{\includegraphics[width=18mm]{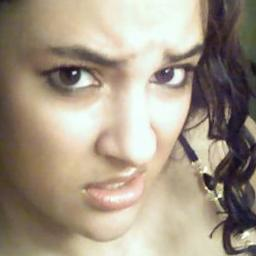}}{\scriptsize{\emph{Angry} (Disgust)}}%
}
\subfloat
{
	\centering
	\stackunder{\includegraphics[width=18mm]{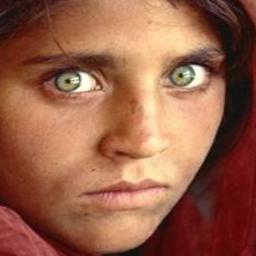}}{\scriptsize{\emph{Neutral} (Fear)}}%
}
\subfloat
{
	\centering
	\stackunder{\includegraphics[width=18mm]{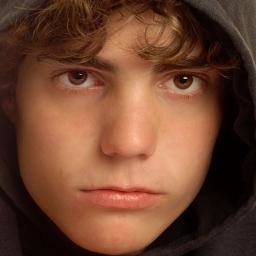}}{\scriptsize{\emph{Angry} (Neutral)}}%
}
\vspace{-0.2cm}
\subfloat
{
	\centering
	\stackunder{\includegraphics[width=18mm]{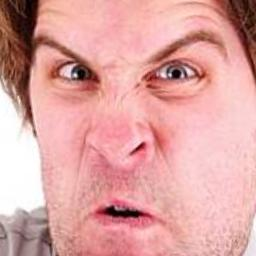}}{\scriptsize{\emph{Disgust} (Angry)}}%
}
\subfloat
{
	\centering
	\stackunder{\includegraphics[width=18mm]{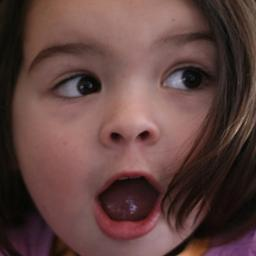}}{\scriptsize{\emph{Happy} (Surprise)}}%
}
\subfloat
{
	\centering
	\stackunder{\includegraphics[width=18mm]{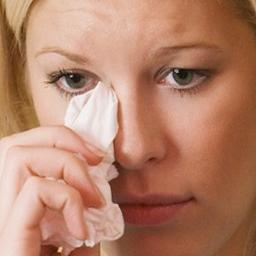}}{\scriptsize{\emph{Neutral} (Sad)}}%
}
\subfloat
{
	\centering
	\stackunder{\includegraphics[width=18mm]{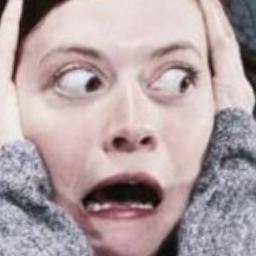}}{\scriptsize{\emph{Surprise} (Fear)}}%
}
\vspace{-0.2cm}
\subfloat
{
	\centering
	\stackunder{\includegraphics[width=18mm]{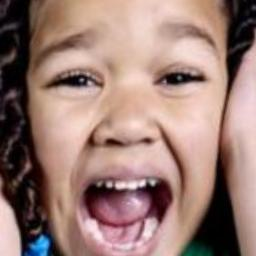}}{\scriptsize{\emph{Happy} (Fear)}}%
}
\subfloat
{
	\centering
	\stackunder{\includegraphics[width=18mm]{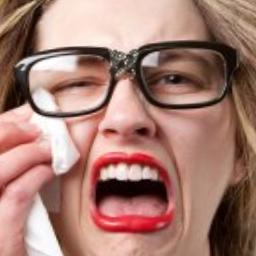}}{\scriptsize{\emph{Angry} (Sad)}}%
}
\subfloat
{
	\centering
	\stackunder{\includegraphics[width=18mm]{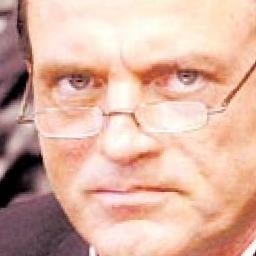}}{\scriptsize{\emph{Neutral} (Angry)}}%
}
\subfloat
{
	\centering
	\stackunder{\includegraphics[width=18mm]{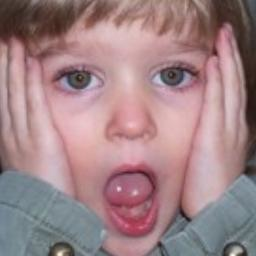}}{\scriptsize{\emph{Neutral} (Surprise)}}%
}
\vspace{0.3cm}
\caption{\label{fig:Sample_MissClassified}
Samples of miss-classified images. Their corresponding ground-truth is given in parentheses.
}
\end{figure}

Figure~\ref{fig:Sample_MissClassified} shows a sample of randomly selected images misclassified by AlexNet trained on the well-labeled and their corresponding ground-truth given in parentheses. 
As the figure shows, it is really difficult to classify some of the images. For example, we were unable to correctly classify the images in the first row. Also, the images in the second row have similarities to the misclassified labels, such as nose wrinkle in disgust, or raised eyebrows in surprise. It should be mentioned that classifying complex facial expressions as discrete emotions, especially in the wild, can be very difficult and even there was only 63.7\% agreement between two human annotators.

\section{Conclusion}
\label{sec:conclusion}

Facial expression recognition in a wild setting is really challenging. Current databases with in wild setting are also either very small or have low resolution without facial landmark points necessary for pre-processing. The Internet is a vast resource of images and it is estimated that over 430 million photos are uploaded on only social network servers every day. Most of these images contain faces, that are captured in uncontrolled settings, illuminations, pose, etc. In fact it is Word \textit{Wild} Web of facial images and it can be a great resource for capturing millions of samples with different subjects, ages, and ethnicity. 

Two neural network architectures were trained in three training scenarios. It is shown that, training on only well-labeled data has higher overall accuracy than training on the mixture of noisy and well-labeled data, even with the noise estimation method. The noise estimation can increase the accuracy in sadness, surprise, fear and disgust expressions, as there were limited samples in well-labeled data. But still training on only well-labeled data has a higher overall accuracy. The reason is that as annotations of web images showed, most of the facial images queried from the web have less than 50\% hit-rates and even for some emotions such as disgust and fear, the majority of the results portrayed other emotions or neutral faces. 

The whole database, query terms, annotated images subset, and their facial landmark points will be publicly available for the research community.

\section{Acknowledgment}
This work is partially supported by the NSF grants IIS-1111568 and CNS-1427872. We gratefully acknowledge the support of NVIDIA Corporation with the donation of the Tesla K40 GPU used for this research.

{\small
\bibliographystyle{ieee}

}

\end{document}